\documentclass[a4paper,UKenglish,cleveref, autoref, thm-restate,authorcolumns]{lipics-v2019}
\title{GeoAI for Knowledge Graph Construction: Identifying Causality Between Cascading Events to Support Environmental Resilience Research} %TODO Please add

\titlerunning{GeoAI for Knowledge Graph Construction} %TODO optional, please use if title is longer than one line

\author{Yuanyuan Tian}{School of Geographical Science and Urban Planning, Arizona State University, Tempe, AZ, USA }{yuanyuantian@asu.edu}{https://orcid.org/0000-0002-5137-1296}{}%TODO mandatory, please use full name; only 1 author per \author macro; first two parameters are mandatory, other parameters can be empty. Please provide at least the name of the affiliation and the country. The full address is optional

\author{Wenwen Li\footnote{corresponding author}}{School of Geographical Science and Urban Planning, Arizona State University, Tempe, AZ, USA }{wenwen@asu.edu}{}{}%TODO mandatory, please use full name; only 1 author per \author macro; first two parameters are mandatory, other parameters can be empty. Please provide at least the name of the affiliation and the country. The full address is optional

\authorrunning{Yuanyuan and Wenwen} %TODO mandatory. First: Use abbreviated first/middle names. Second (only in severe cases): Use first author plus 'et al.'

\Copyright{Yuanyuan and Wenwen} %TODO mandatory, please use full first names. LIPIcs license is "CC-BY";  http://creativecommons.org/licenses/by/3.0/

 %TODO mandatory; please add comma-separated list of keywords

\category{} %optional, e.g. invited paper

\relatedversion{} %optional, e.g. full version hosted on arXiv, HAL, or other respository/website
\supplement{}%optional, e.g. related research data, source code, ... hosted on a repository like zenodo, figshare, GitHub, ...

\funding{National Science Foundation (NSF) under award OIA-2033521.}%optional, to capture a funding statement, which applies to all authors. Please enter author specific funding statements as fifth argument of the \author macro.

\acknowledgements{Thank Mark Schildhauer and Rui Zhu at University of California Santa Barbara, and Cogan Shimizu at Kansas State University for ontology engineering and comments. }%optional

\nolinenumbers %uncomment to disable line numbering

\EventLongTitle{1st International Workshop on
Methods, Models, and Resources for Geospatial Knowledge Graphs and GeoAI}
\EventShortTitle{GeoKG\&GeoAI}
\EventAcronym{GeoKG \& GeoAI}
\EventYear{2021}
\EventDate{September 27, 2021}
\EventLocation{Poznań, Poland}
\EventLogo{}
\ArticleNo{06}
\begin{document}

\maketitle

\begin{abstract} Knowledge graph technology is considered a powerful and semantically enabled solution to link entities, allowing users to derive new knowledge by reasoning data according to various types of reasoning rules. However, in building such a knowledge graph, events modeling, such as that of disasters, is often limited to single, isolated events. The linkages among cascading events are often missing in existing knowledge graphs. This paper introduces our GeoAI (Geospatial Artificial Intelligence) solutions to identify causality among events, in particular, disaster events, based on a set of spatially and temporally-enabled semantic rules. Through a use case of causal disaster events modeling, we demonstrated how these defined rules, including theme-based identification of correlated events,  spatiotemporal co-occurrence constraint, and text mining of event metadata, enable the automatic extraction of causal relationships between different events. Our solution enriches the event knowledge base and allows for the exploration of linked cascading events in large knowledge graphs, therefore empowering knowledge query and discovery.\end{abstract}

\vspace{-0.2cm}
\section{Introduction}
\label{sec:Introduction}
\vspace{-0.3cm}
Global warming and climate change exacerbate the frequency and severity of global disasters. Since 1980, nearly 300 weather and climate disaster events have hit the U.S., each causing over 1 billion dollars of economic loss, and the National Ocean and Atmospheric Agency (NOAA) released a recent Billion-Dollar Disaster Report showing 2020 as a historic year of extreme climate disasters \cite{smith2020us}. There is an urgent need to improve our understanding of both historical and incoming disastrous events to develop disaster preparation and response solutions to increase our urban and ecosystem's resilience to these events \cite{marchese2018resilience}. 

In recent years, knowledge graphs (KG) have become a popular approach for building machine-understandable semantic models of entities and their interrelationships. These entities could be a static object, such as a building or an event, such as a hurricane. Building a large knowledge graph with linked data can enable a new paradigm for question answering and knowledge discovery \cite {li2019ontology}. In disaster modeling, knowledge graphs also play a key role in integrating multi-modal data for improved disaster management and response \cite{fan2021disaster}. Since disasters often have a cause-and-effect relationship that reflects the interconnections between events, linking cascading events has become a central task in the semantic and formal modeling of disasters. However, existing efforts have been mostly centered on modeling single events, and the interrelations among them which are substantially missing \cite{pescaroli2015definition}. The entity alignment technique has the potential to discover identical or related geospatial entities and harmonize heterogeneous datasets\cite{sun2020aligning}, yet barely deliver the interlinking value of cascading disaster events. 
 
This paper aims to address the limitation in knowledge graph modeling of cascading events, by developing a GeoAI \cite{li2020geoai} solution with a set of semantic rules for causality identification. These rules build upon each other and help to derive from (1) thematically correlated events (i.e., those clustered in a single large event collection) to (2) spatially and temporally co-occurred events (i.e., cascading events that often occur in sequence and are spatially correlated), to (3) semantically filtered events by mining the events metadata to further determine their causal relationship. We use disaster modeling as our use case to demonstrate the effectiveness of the proposed approach. 

\vspace{-0.4cm}
\section{Methodology}
\label{sec:Methodology}
\vspace{-0.2cm}
We use severe storms and extreme events data to build a knowledge graph (KG) that allows for the exploration of cascading severe weather events. The goal is to create linkages in a disaster domain KG to allow semantic query and knowledge inference. Three semantic rules are defined to identify causality among events and answer pilot questions, such as “What are cascading E1 and E2 severe events during a specific disaster D? ” In this query, E1 and E2 are words related to event types, D is the name of a named disaster. Rule \#1 identifies thematically correlated events, rule \#2 finds disaster events that match with report information according to spatial and temporal co-occurrence constraints, and rule \#3 uses metadata text mining to determine the causal relationships between events. 

\vspace{-0.4cm}
\subsection{Data and Graph Building}
\vspace{-0.2cm}
The Storm Event Database covers the U.S. from January 1950 to April 2021 and contains event reports collected by National Weather Service (NWS). This database is used as our test dataset, in which an episode is an entire storm system and includes various types of events. An event represents a meteorological event leading to fatalities, injuries, and other kinds of damages \cite{murphy2016storm}. Multiple episodes can be related to a disaster such as Hurricane Sandy, and each episode can encompass multiple severe events. There are 48 different types of meteorological events, from localized thunderstorms, tornadoes, and flash floods to regional events such as hurricanes and winter storms. Each record documents event attributes related to an event type, event location, event start and end time, weather conditions, and event and episode narratives. Impacts also include crop and property damage as well as direct and indirect death and injury. A disaster ontology is constructed using the original data structure and Sensor-Observation-Sampling-Actuator ontology to achieve an explicit hierarchy. Then, the storm event records are populated into the ontology to build the KG.

\vspace{-0.4cm}
\subsection{Thematically Correlated Event Identification}
\vspace{-0.2cm}
To deconstruct the pilot question, the first step is to understand the E1 and E2 in the question. That human-readable question should be translated to a general term rather than the explicit event type named E1 such as “flood” because there are other types such as “flash flood” which also satisfy the question's semantics. Even domain experts may not be familiar with this dataset’s event type classification standard. Thus, we introduce the concept of theme to interpret E1 and E2 as “E1-themed” and “E2-themed" correlated event types, respectively. For example, “hurricane-themed” and “flood-themed” events are defined as: “Flood-themed” event types - Flood, FlashFlood, CoastalFlood, LakeshoreFlood; “Hurricane-themed” event types - Hurricane, Heavy Rain, High Surf, Marine Hurricane Typhoon, Marine Tropical Depression, Marine Tropical Storm, Sneakerwave, StormSurgeTide, Tropical Depression, Tropical Storm. Through this query expansion method, all thematically relevant events can be retrieved from the dataset. 

\vspace{-0.4cm}
\subsection{Disaster Matching According to Spatial and Temporal Co-occurrence Constraint}
\vspace{-0.2cm}
The next rule is to support the identification of severe events related to a disaster named D. This is challenging due to the data structure. NWS assigned a unique numerical ID to an episode or an event; however, the disaster name was not specified as an attribute. We proposed a solution to match disasters with events based on two constraints. First, we used a keyword to match the disaster name with the episode or event narrative to identify the data records containing the needed disaster information. However, matching by disaster name is insufficient because some narratives include comparison description sentences referring to a previous disaster. Thus, we added the second constraint based on spatial and temporal co-occurrences of the known events. The event geometry or occurring location should overlap with the regions that have reported impacts during a given time period. The event reports had timestamps with a duration measured to the minute, indicating that the events had short-term impacts. Since some impacts can occur before the event's formed date or after the dissipated date, a seven-day time buffer is added to the duration coverage.

\vspace{-0.4cm}
\subsection{Metadata Text Mining}
\vspace{-0.2cm}
Metadata text mining is used to find causality among "E1--themed" and "E2--themed" event narratives to empower linkage discovery and prediction. The first task is to build a knowledge base containing terms that represent causality and extract connected events. A thesaurus is built by integrating causal keywords and phrases such as "lead to", "result in'', and "induce" collected from Merriam-Webster Dictionary, Lexico, and \cite{aghakouchak2018natural} as connector candidates. 
% <reconstruct the rest of the section accordingly>
The next task is to construct event causality networks from event narratives. Specific casual events are extracted and linked, then a hierarchical causality generation method \cite{zhao2017constructing} is used to uncover the high-level causality patterns on top of the specific one.
% [cause, beget, breed, bring, bring about, bring on, catalyze, create, do, draw on, effect, effectuate, engender, generate, induce, invoke, make, occasion, produce, prompt, result, spawn, translate, work, yield, bring about, give rise to, be the cause of, lead to, result in, create, begin, produce, generate, originate, engender, spawn, occasion, effect, bring to pass, bring on, precipitate, prompt, provoke, kindle, trigger, make happen, spark off, touch off, stir up, whip up, induce, inspire, promote, foster, couple, connect, chain, link, impact]
% Set. Candidate word set of causality

\vspace{-0.4cm}
\begin{figure}[h!]
\begin{center}
  \includegraphics[width=1\textwidth]{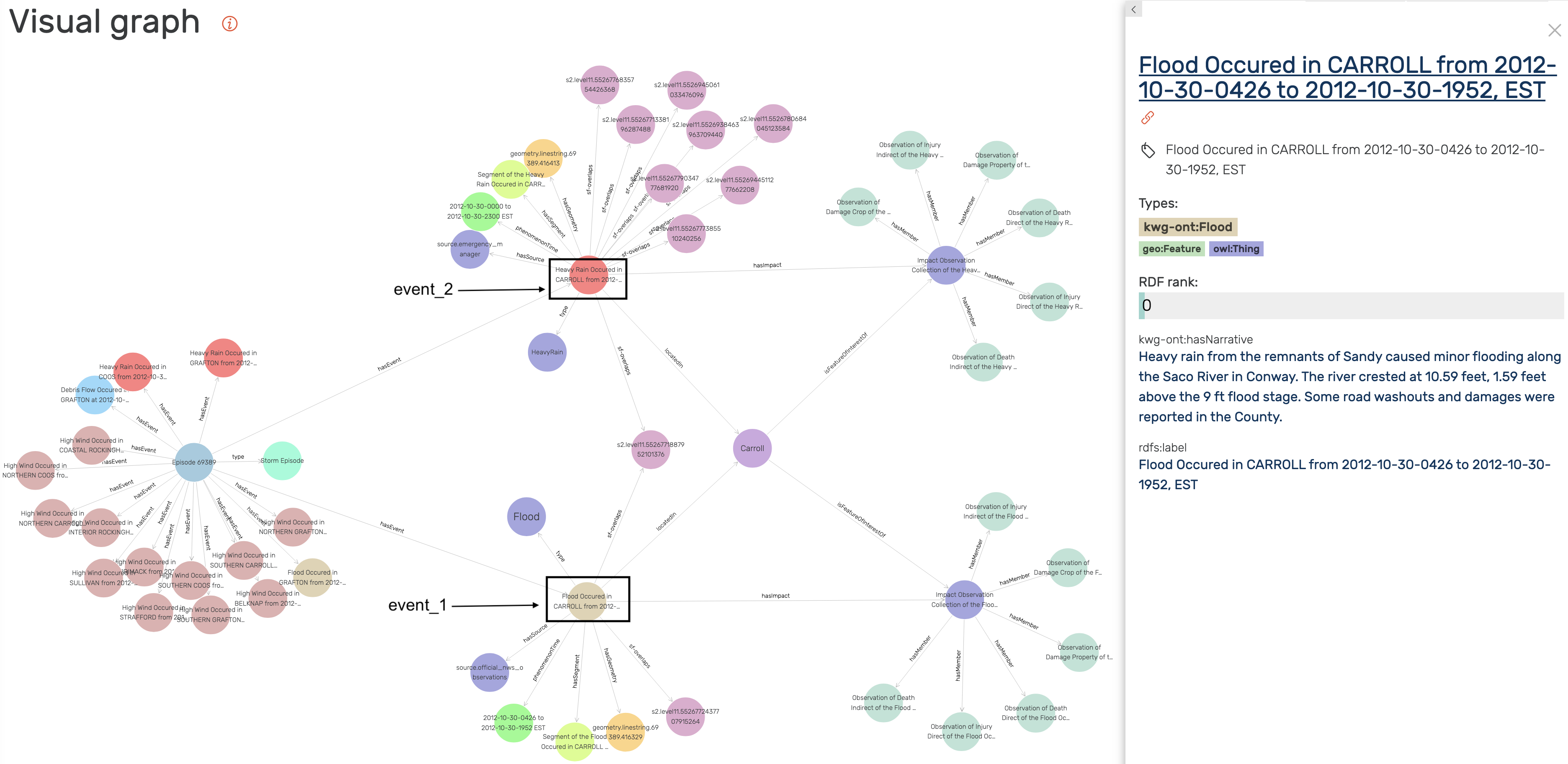}
\vspace{-0.4cm}
\end{center}
   \caption{A pair of linked events: event 1, a flood and event 2, heavy rain.}
\label{fig:Figure2}
\end{figure}

\vspace{-0.65cm}
\section{Result}
\label{sec:Results}
\vspace{-0.3cm}
According to the semantically defined rules, the event network that has causal relationships can be found. For example, “What are cascading flood and hurricane events during Hurricane Sandy?” 
%>> I suggest to remove this below sentence. 
% As expected, the number of severe events decreases when adding more rules, indicating that the rules-based constraints refine results and remove events without strong causal relationships.
Figure~\ref{fig:Figure2} depicts findings of a pair of occurrences of events from Hurricane Sandy's devastation. A beige node is a flooding event and the red node refers to a heavy rain event. Each resultant event is depicted by time, location, and damage caused. The flood narrative shows the correct name for Sandy and claims that the heavy rains caused minor flooding. %<double check this last sentence)>. If heavy rain caused by a future hurricane occurs in Carroll again, the KG can tell people probably there will be a cascading event such as a flood.

% \begin{figure}[h!]
% \begin{center}
%   \includegraphics[width=0.95\textwidth]{Figure1.pdf}
% \end{center}
%   \caption{Severe events related to Hurricane Katrina.}
% \label{fig:Figure1}
% \end{figure}

\vspace{-0.45cm}
\section{Conclusions and Future Work}
\label{sec:Conclusions and Future Work}
\vspace{-0.3cm}
This paper develops a new GeoAI based semantic analysis approach that connects the cascading disaster events. The method is capable of discovering the interconnections between many types of severe weather events, their effects on the surrounding environment, and impacts on human society. Leveraging the information about interlinked events, researchers and decision makers could gain a better understanding of compound hazards and cascading events for disaster preparation and response. This will increase urban and environmental resilience to incoming disasters. In the future, we will optimize the spatial and temporal co-occurrence constraint by testing different topological measures and temporal buffers. Also, refine the methodology to include deep learning based link prediction for finding the causal relationships to establish the semantically enriched linked data in an automatic manner. We will also extend the methodological framework to support the construction of a large-scale knowledge graph with multiple types of disasters, and add quantitative evaluation results.

%%
%% Bibliography
%%

%% Please use bibtex, 
\vspace{-0.5cm}
\bibliography{lipics-v2019-sample-article}

\end{document}